\documentclass[a4paper,fleqn]{cas-dc}

\usepackage[numbers]{natbib}

\usepackage{makecell}

\usepackage{multirow}
\usepackage{colortbl}
\usepackage{subfigure}  
\usepackage{subcaption}
\usepackage{svg}
\usepackage{float}

\usepackage{algorithm}
\usepackage{algorithmic}

\usepackage{soul} 
\usepackage{color, xcolor} 
\soulregister{\cite}7 
\soulregister{\citep}7 
\soulregister{\citet}7 
\soulregister{\ref}7 
\soulregister{\pageref}7 

\usepackage{bbding}

\usepackage{enumitem}

\usepackage{amsmath,amssymb}                
\usepackage{array}
\usepackage{bm}
\usepackage{booktabs} 
\usepackage{changepage}
\usepackage{eqparbox}
\usepackage{mdwmath}
\usepackage{mdwtab}
\usepackage{url}

\colorlet{shadecolor}{yellow}

\usepackage{graphicx}
\graphicspath{{../pdf/}{../jpeg/}}
\DeclareGraphicsExtensions{.pdf,.jpeg,.png,.svg}

\def\tsc#1{\csdef{#1}{\textsc{\lowercase{#1}}\xspace}}
\tsc{WGM}
\tsc{QE}

\begin{document}
\sloppy
\let\WriteBookmarks\relax
\def\floatpagepagefraction{1}
\def\textpagefraction{.001}

\shorttitle{M. Huang et~al. Knowledge-Based Systems}    

\shortauthors{M. Huang et~al.} 

\title [mode = title]{
GRIN+: Towards Fast Yet Effective Machine Unlearning for Imbalanced
Medical Data 
}  



%


\author[1]{Minghui Huang}
\ead{2112433114@e.gzhu.edu.cn}








\author[1]{Junxiao Wang}
\ead{junxiao.wang@gzhu.edu.cn}
\cormark[1]

\affiliation[1]{organization={Guangzhou University}, 
                country={China},
                }
                
\cortext[1]{Junxiao Wang is the corresponding author}















\begin{abstract}
As deep learning models become fundamental to modern healthcare, the ``Right to be Forgotten'' mandated by privacy regulations like GDPR and HIPAA necessitates effective machine unlearning (MU) to remove sensitive patient data from trained models. However, existing MU techniques often struggle with a fundamental ``privacy-efficiency-utility'' (PEU) trilemma, particularly in medical scenarios where data is frequently characterized by severe class imbalance and long-tailed distributions. In such cases, standard unlearning methods can fail to protect key clinical knowledge or mistakenly delete features essential for diagnosing rare conditions due to the gradient dominance of majority classes. To address these challenges, we propose GRIN+, a novel machine unlearning framework designed for fast and precise data erasure in imbalanced medical scenarios. GRIN+ decouples unlearning-specific knowledge from generalized representations at the parameter level by analyzing the gradient contributions of both ``forget'' and ``retain'' sets. It introduces a class-adaptive influence scoring mechanism to rectify gradient dominance and employs a direction-constrained update strategy to prevent the unintended erosion of vital clinical knowledge. Comprehensive benchmarking across multiple medical datasets, including skin cancer (ISIC), brain tumor (MRI), and breast ultrasound (BUSI), demonstrates that GRIN+ achieves an optimal balance of the PEU trilemma. Experimental results show that GRIN+ maintains high diagnostic accuracy and robust privacy while significantly enhancing runtime efficiency compared to existing baselines. We open-source the GRIN+ code and benchmarks to support further research.
\end{abstract}


\maketitle

\sloppy
\section{INTRODUCTION}
\textbf{Background.}
As medical AI rapidly enters clinical use, deep learning models have become a cornerstone of modern healthcare~\cite{esteva2019guide}. They facilitate critical tasks such as medical image analysis (including MRI~\cite{jyothi2023deep}, ultrasound~\cite{cao2019experimental}, and skin screening~\cite{dildar2021skin}), early disease detection~\cite{wang2020early}, and clinical decision-making~\cite{adlung2021machine}. 
Since these models are trained on sensitive patient data, including images and clinical metadata, they are subject to strict international privacy laws. For instance, the GDPR~\cite{gdrp} and HIPAA~\cite{hipaa} grant patients the ``Right to be Forgotten'', allowing them to request the total removal of their data. This creates a significant hurdle for deep learning: removing data from a database does not erase its impact on a trained model.

Studies~\cite{carlini2019secret} have reported that models often retain traces of individual training samples within their architecture. In a medical context, this residual data can be targeted by membership inference attacks~\cite{carlini2022membership}, risking the exposure of private patient information and leading to serious legal and clinical consequences.

Machine unlearning~\cite{bourtoule2021machine} has been proposed as a solution to these challenges, aiming to precisely and efficiently remove the influence of specific training data from pre-trained models. Existing techniques generally fall into two categories: exact unlearning~\cite{guo2023certifieddataremovalmachine} and approximate unlearning~\cite{izzo2021approximatedatadeletionmachine}. Exact unlearning focuses on providing certified deletion or provable guarantees. Within this category, retraining the model from scratch on the retained data is considered the gold standard~\cite{thudi2022necessity}. Nevertheless, such retraining-based methods incur excessive computational costs. This makes them increasingly impractical for clinical settings that demand low latency and frequent model updates~\cite{kumar2025integrating}.

Alternatively, approximate machine unlearning offers a practical way to improve efficiency. These methods avoid the burden of full retraining by fine-tuning model parameters, utilizing techniques like influence functions~\cite{koh2017understanding} or Fisher Information Matrices (FIM)~\cite{martens2020new} to estimate parameter importance. While they have made impressive contributions in standard computer vision tasks, current validation remains largely confined to vanilla datasets like CIFAR~\cite{shi2024deepclean,parameter_editing:golatkar2020eternal,warnecke2023machine,wang2022federated}. This suggests that their efficiency gains typically rely on the implicit assumption that the underlying training data is class-balanced.

\begin{figure*}[htbp]
  \setlength{\abovecaptionskip}{-0.1cm}
  \centering
  \includegraphics[width=1.0\textwidth]{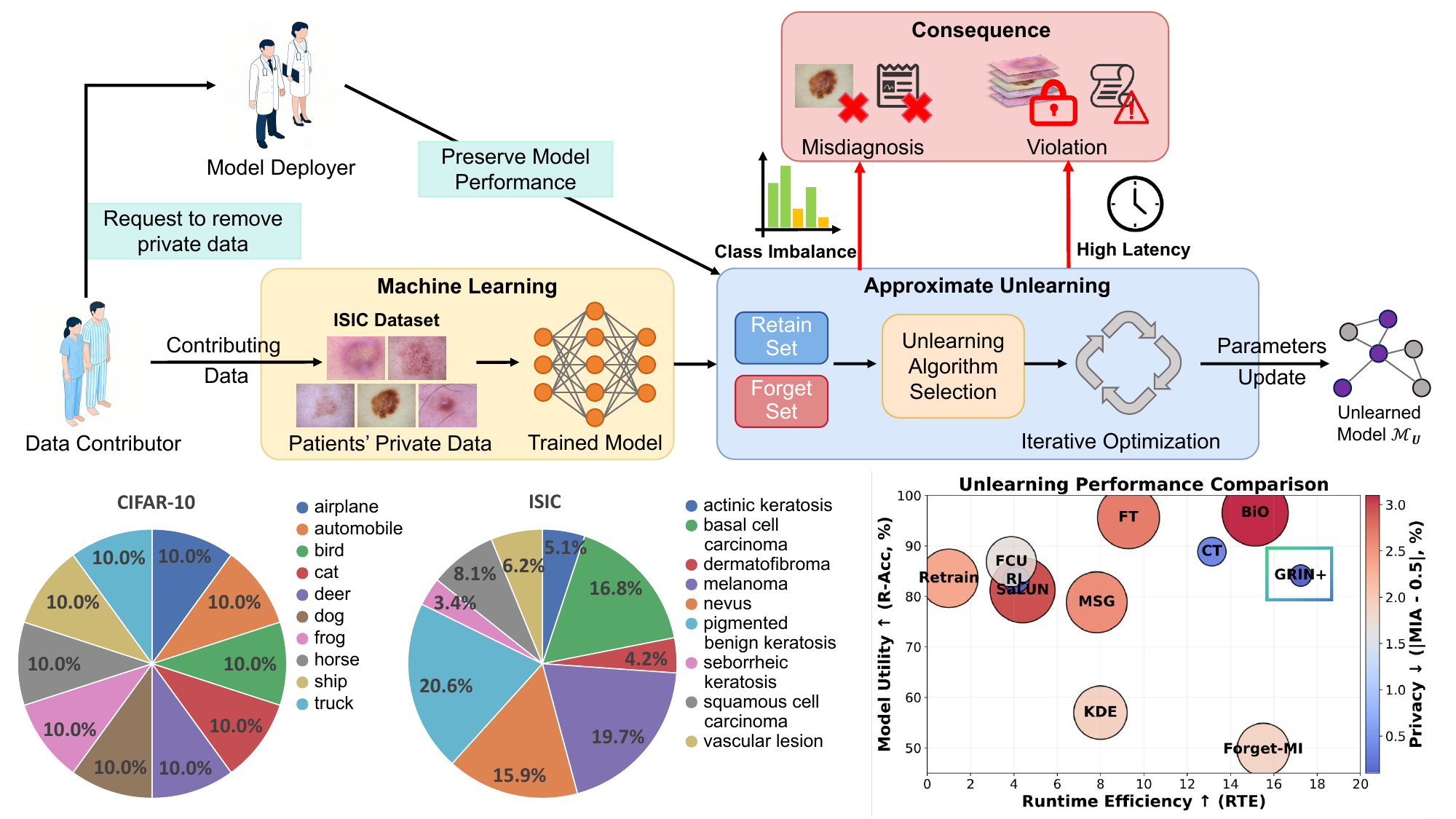}\\
  \caption{Limitations of approximate machine unlearning in medical scenarios. \textbf{Top:} Existing methods rely on uniform parameter updates and iterative optimization, leading to two critical issues: (1) gradient dominance from majority classes in imbalanced data suppresses minority representations, causing degraded recognition of rare diseases (Misdiagnosis); (2) high computational overhead results in significant latency, hindering timely privacy compliance (Violation). \textbf{Bottom:} Left: Class distribution comparison between CIFAR-10 and the highly imbalanced ISIC dataset. Right: Unlearning performance comparison. RTE (x-axis, higher is better) vs. R-Acc (y-axis, higher is better); bubble size directly reflects $|\text{MIA} - 0.5|$ (the smaller and lighter color the bubble, the better the unlearning effectiveness and the lower the privacy risk). Top-right with small and light-colored bubbles indicates fast and effective unlearning.}\label{motivation}
\end{figure*}

\textbf{Motivation.}
The assumption of balanced data, however, is often violated in medical imaging tasks. Clinical data in the real world usually follows a long-tailed distribution characterized by severe class imbalance (as shown in the lower left of Figure~\ref{motivation}). In these cases, minority classes (like rare diseases) have very few samples but carry much greater clinical importance~\cite{pan2025long}. During training, gradients are dominated by majority-class samples, which weakens the model's ability to recognize rare but critical pathological patterns. Benchmarks such as CXR-LT~\cite{lin2025cxr} have demonstrated that standard models tend to prioritize head-class performance while neglecting the tail classes, a problem that intensifies during the unlearning process~\cite{yu2026falw}. Efficiency-oriented methods like influence functions and FIM typically rely on the implicit assumption of class balance~\cite{koh2017understanding,martens2020new}. This causes their parameter importance estimates to skew toward majority classes. On imbalanced medical data, these methods fail to protect key clinical knowledge and may mistakenly delete features essential for diagnosing rare conditions. Consequently, existing unlearning solutions cannot fully meet clinical requirements, while maintaining efficiency and reliability in long-tailed medical scenarios, as illustrated in the lower right of Figure~\ref{motivation}.

\textbf{As our first contribution,} 
we identify a fundamental ``privacy-efficiency-utility'' (PEU) trilemma inherent in machine unlearning for imbalanced medical data. Specifically, an ideal unlearning framework must fulfill three core requirements: (1) Privacy: the unlearned model must not leak membership information from the ``forget set'', thereby complying with stringent data protection laws; (2) Efficiency: the unlearning process must be sufficiently rapid to support clinical environments that demand low latency and frequent model updates; and (3) Utility: the resulting model must retain high diagnostic performance, even when navigating the complexities of imbalanced medical data.

\textbf{Our second contribution:} 
Motivated by the aforementioned trilemma, we propose GRIN+, a novel machine unlearning framework designed for fast and effective data erasure in imbalanced medical scenarios. GRIN+ explicitly decouples unlearning-specific knowledge from generalized representation capabilities at the parameter level. By jointly analyzing the gradient contributions of both the ``forget'' and ``retain'' sets, the framework quantifies parameter-wise influence to pinpoint a sparse subset closely associated with the target data, thereby mitigating parameter misattribution. Furthermore, GRIN+ introduces a class-adaptive influence scoring mechanism to rectify the gradient dominance of majority classes inherent in long-tailed distributions. Integrated with a direction-constrained update strategy, this mechanism prevents the unintended erosion of vital clinical knowledge during parameter adjustment. Ultimately, by executing directional updates based on gradient disparities, GRIN+ precisely eliminates target information while preserving stable diagnostic performance.

\textbf{As our third contribution,}
we comprehensively benchmark GRIN+ across multiple medical datasets, and the results demonstrate that GRIN+ achieves an optimal balance between unlearning efficiency, model utility, and privacy. Specifically, the resulting models maintain high accuracy on both the retain and test sets. Notably, the Membership Inference Attack (MIA) score reaches 49.90\%, nearly equivalent to a random guess, indicating robust privacy preservation. Furthermore, GRIN+ significantly enhances runtime efficiency compared to baselines. Finally, comprehensive ablation studies confirm the indispensable contribution of each individual component to the overall performance.


This paper can be summarized as three-fold: 

\begin{itemize}
    \item \textbf{Problem.} 
    We first reveal a ``privacy-efficiency-utility'' (PEU) trilemma in machine unlearning for imbalanced medical data, requiring a PEU instance to simultaneously achieve privacy (preventing data leakage from the forget set), efficiency (ensuring rapid updates for clinical use), and utility (maintaining high diagnostic accuracy despite data imbalances).
\end{itemize}

\begin{itemize}
    \item \textbf{Method.} 
    To address the PEU trilemma, we propose GRIN+, an unlearning framework that achieves fast, precise data erasure in imbalanced medical scenarios by decoupling unlearning-specific knowledge from general representations at the parameter level, utilizing a sparse subset selection based on gradient analysis of ``forget'' and ``retain'' sets, and employing a class-adaptive influence scoring mechanism with direction-constrained updates to prevent majority-class dominance and preserve vital clinical performance.
\end{itemize}

\begin{itemize}
    \item \textbf{Results.} We conduct comprehensive benchmarks and demonstrate that GRIN+ achieves an optimal balance of the PEU trilemma by maintaining high diagnostic accuracy, ensuring robust privacy, and significantly improving runtime efficiency.
    
\end{itemize}

\textbf{Open-source.} We release the GRIN+ code\footnote{https://github.com/gzhu-hcai/Med-Unlearn} to the community for further research.

\sloppy
\section{Related Work}

\subsection {Machine Unlearning}
Machine unlearning~\cite{bourtoule2021machine} aims to precisely and efficiently remove the influence of specific training data from pre-trained models. Existing methods are mainly divided into exact unlearning~\cite{guo2023certifieddataremovalmachine} and approximate unlearning~\cite{izzo2021approximatedatadeletionmachine}.

While exact unlearning provides strict data deletion guarantees, early methods like SISA~\cite{bourtoule2021machine}, which use data sharding and submodel retraining, face significant hurdles. They disrupt standard training pipelines and impose high storage and computational costs, making them difficult to scale for medical AI. Consequently, approximate unlearning has become the primary focus for improving efficiency. These techniques use tailored loss functions to update parameters so that the model mimics a retrained state while protecting against membership inference~\cite{carlini2022membership} and reconstruction attacks~\cite{balle2022reconstructing}. Representative approaches include CF-K~\cite{goel2022towards} (freezing shallow layers) and SCRUB~\cite{kurmanji2023towards} (using distillation and adversarial learning). Although more efficient than retraining, these methods still rely on global fine-tuning, which can be costly. Moreover, managing the delicate balance between forgetting and retaining knowledge remains a challenge, often leading to a trade-off where either unlearning is incomplete or the model's accuracy on the remaining data is compromised.

\subsection {Unlearning in Medical Context}
Closest to our research are strategies that involves parameter importance estimation followed by selective updates, another mainstream paradigm in approximate unlearning. These methods typically leverage influence functions (approximating the inverse Hessian)~\cite{koh2017understanding,warnecke2023machine} or the Fisher Information Matrix (FIM)~\cite{martens2020new,parameter_editing:golatkar2020eternal,shi2024deepclean} to quantify the impact of individual training samples on model predictions.
Nevertheless, these techniques face significant hurdles. First, calculating inverse Hessian or Hessian-vector products is still too expensive for large-scale deep models. Second, influence functions are unstable in deep, non-convex landscapes, especially when dealing with a nearly singular Hessian~\cite{basu2021influence}. In medical contexts where data is long-tailed, rare diseases have few samples and low curvature, making the importance estimates highly unreliable. Finally, because FIM is calculated as an expectation over the entire data distribution, it is easily biased by class imbalance. Since head classes dominate the gradients, the importance of majority-class features is overestimated. Consequently, the model may mistakenly delete information vital for recognizing rare pathologies.

Research specifically tailored to medical scenarios remains limited and somewhat fragmented. While some approaches utilize low-rank adaptation or distillation for selective erasure~\cite{datta2025erase}, or adjust decision boundaries through bilevel optimization~\cite{nahass2025targeted}, these methods often rely on auxiliary constraints or teacher models, making them vulnerable to the unstable boundaries inherent in imbalanced data. Other studies remain confined to niche tasks, such as reconstruction~\cite{xue2024erase}, or specific frameworks like federated~\cite{other_med:deng2024enable} and multimodal unlearning~\cite{other_med:hardan2025forget}, which restricts their general applicability. 
In summary, machine unlearning in medical imaging continues to face two formidable challenges: biased parameter importance estimation under long-tailed distributions, and the persistent difficulty in reconciling efficiency with model utility.

\section{Preliminaries}

We formalize the machine unlearning problem in the medical domain as follows:

Given a training set $\mathcal{D}_{\text{train}} = \{(x_i, y_i)\}_{i=1}^N$ and a pre-trained model $\mathcal{M}$ (called the original model) with parameters $\theta^0$, the core goal of machine unlearning (MU) is to eliminate the influence of a specific subset $\mathcal{D}_f = \{(x_j^{(f)}, y_j^{(f)})\}_{j=1}^M \subseteq \mathcal{D}_{\text{train}}$. This subset, defined as the \textit{forget set}, corresponds to a single patient's record or a small batch of patient data, with $M \ll N$ holding true. The remaining part of the training set, $\mathcal{D}_r = \mathcal{D}_{\text{train}} \setminus \mathcal{D}_f$, is named the \textit{retain set}. 
In a medical context, $\mathcal{D}_r$ statistically reflects the knowledge distribution the model needs to retain.

These two sets are disjoint and complementary components of the original training set. Applying MU yields an unlearned model $\mathcal{M}_U$. Its objective is to achieve performance comparable to a model retrained from scratch on $\mathcal{D}_r$. 
We use $\theta^0$ and $\theta_r$ to represent the weights of the original and retrained models, respectively. For evaluation, we introduce two held-out sets: the validation set $\mathcal{D}_V$ and the test set $\mathcal{D}_T$. Both sets follow the same distribution as $\mathcal{D}_{\text{train}}$. We regard the retrained model's accuracy as the optimal standard. A key assumption here is that the MU method can access $\theta^0$, $\mathcal{D}_r$, and $\mathcal{D}_V$. The main symbols used throughout this paper are summarized in Table~\ref{tab:notation}.

\begin{table}  
  \centering
  \caption{Summary of main notations.}
  \label{tab:notation}
  \begin{tabular}{ccc}
    \toprule
    \midrule
    Symbol & Description \\
    \midrule
    $\mathcal{D}_{\text{train}}$	& Full training dataset \\
    $\mathcal{D}_r$	& Retain (remaining) dataset \\
    $\mathcal{D}_f$	& Forget dataset \\
    $(x, y)$ & Input sample and corresponding label \\
    $\mathcal{M}_U$	& Unlearned model \\
    $\boldsymbol{\theta}^0$ & Original trained model parameters \\
    $\mathcal{L}(\cdot)$ & Loss function \\
    $\nabla_{\boldsymbol{\theta}}\mathcal{L}$ & Gradient of loss w.r.t. parameters \\
    $\bm{G}_f$ & Gradient computed on forget set \\
    $\bm{G}_r$ & Gradient computed on retain set \\
    $S_i$ & Gradient influence score for parameter $i$ \\
    $p$ & Selection ratio \\
    $|\Theta|$ & Total number of parameters \\
    $\delta_i$ & Parameter update perturbation \\
    $\|\cdot\|$ & Norm operator (default: $\ell_2$ norm) \\
    $\odot$ & Element-wise multiplication \\
    \midrule
  \end{tabular}
\end{table}

\begin{figure*}[htbp]
  \setlength{\abovecaptionskip}{-0.1cm}
  \centering
  \includegraphics[width=1.0\textwidth]{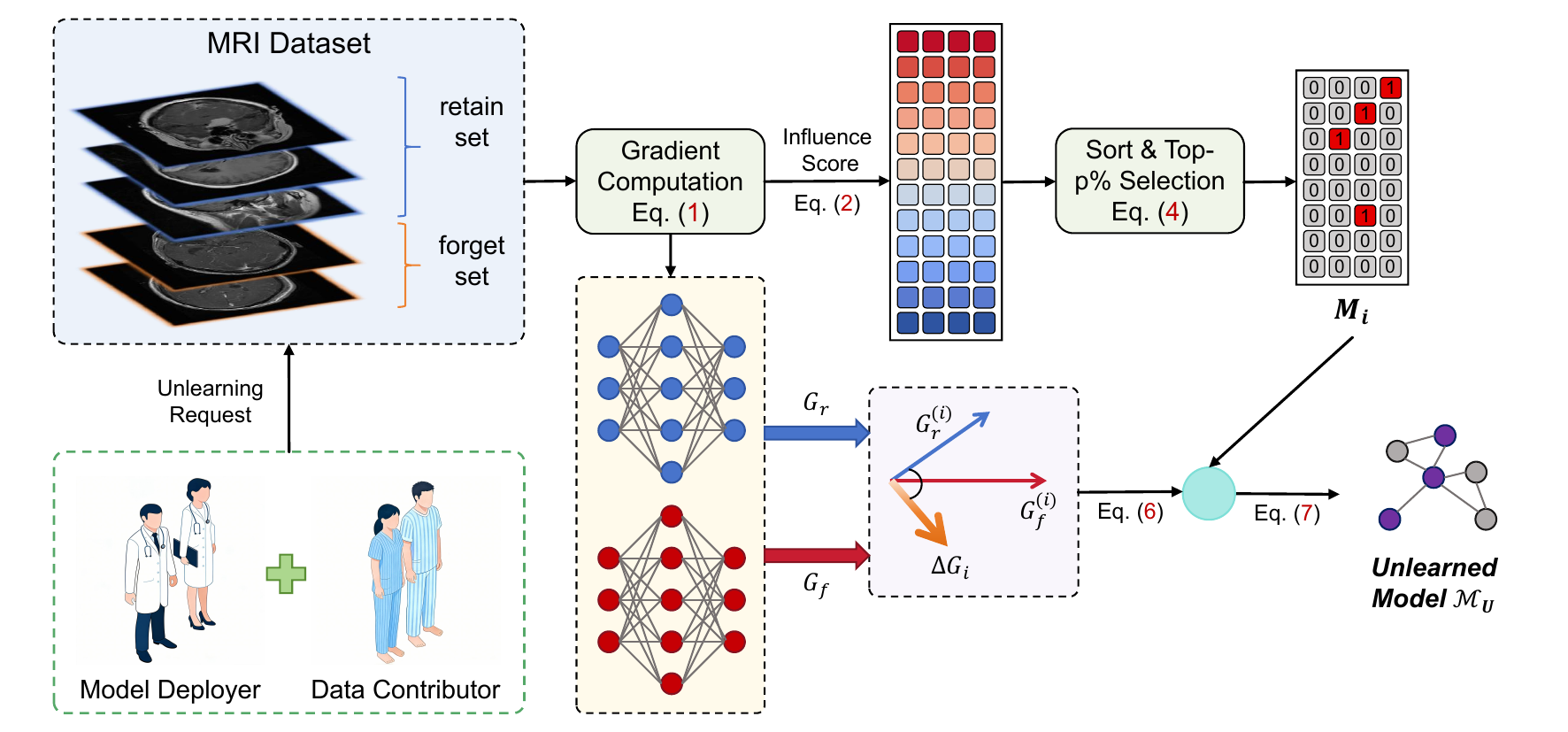}
  \caption{Overview of the proposed GRIN+ framework. After a patient requests the removal of relevant private data, the framework first partitions the data into retain and forget sets. It then estimates parameter-wise influence using gradient-based scores to identify key parameters. Finally, constrained perturbations are applied to these parameters to update the model, enabling selective knowledge removal while preserving performance on the retained data.}\label{framework}
\end{figure*}

\section{Methodology}
The core of our method GRIN+\footnote{Our method's name ``GRIN+'' is the abbr. of ``\underline{G}radient \underline{R}atio-Based Adaptive \underline{IN}fluence
Estimation \underline{+} Regularization-Constrained and Directed
Parameter Perturbation''.} is that, given a pre-trained model, we quantify the relative importance of each parameter for the forgetting operation by analyzing the gradient differences of the loss function on the forget set and a representative retain set. Based on this, we apply only small, constrained perturbations in a specific direction to a filtered subset of critical parameters, thereby effectively `erasing' the memory of target samples while preserving the stability of the model's overall performance. The overview of the pipeline is shown in Figure~\ref{framework}.

\subsection{Gradient Ratio-Based Adaptive Influence Estimation}
We first compute the loss function gradients of the model with respect to $D_f$ and $D_r$ at the current parameters:
\begin{equation}\label{gradients}
\begin{array}{l}
    \bm{G}_f = \nabla_\theta \mathcal{L}(D_f; \theta), \ \ \ \
    \bm{G}_r = \nabla_\theta \mathcal{L}(D_r; \theta),
\end{array}
\end{equation}
where $\mathcal{L}$ is the model's loss function (e.g., cross-entropy loss), and $\bm{G}_f$, $\bm{G}_r$ are gradient vectors with the same dimensionality as the parameters.

Based on the intuition that ``ideal forget parameters should have large gradients on $D_f$ (sensitive and easily modified) and small gradients on $D_r$ (harmless to modify)'', we define a base influence score for each parameter $\theta_i$. To address the prevalent class imbalance problem in medical data, we introduce a class balancing factor for adaptive adjustment:
\begin{equation}\label{balancing_factor}
    \begin{array}{l}
    s_i^{\text{base}} = \frac{|\bm{G}_{f,i}|}{|\bm{G}_{r,i}| + \varepsilon}, \ \ \ \
    S_i = s_i^{\text{base}} \cdot w_c.
    \end{array}
\end{equation}

Here, $|\bm{G}_{f,i}|$ and $|\bm{G}_{r,i}|$ represent the magnitudes of the $i$-th parameter dimension components of gradients $\bm{G}_f$ and $\bm{G}_r$, respectively. $\varepsilon$ is a small positive number for numerical stability, set to the $k$-th percentile of the sequence of absolute values of all components of $\bm{G}_f$ (empirically $k=5$). $w_c$ is the weight of class $c$ to which the forget samples belong, defined as:
\begin{equation}
   w_c = \frac{N}{C \cdot n_c}, 
\end{equation}
where $N$ is the total number of samples in the forget set $\mathcal{D}_f$, $C$ is the total number of classes, and $n_c$ is the number of samples of class $c$. This weight assigns higher weights to minority class samples and lower weights to majority class samples, thereby preventing the forgetting operation from excessively damaging the model's recognition capability for rare diseases or uncommon cases (minority classes) due to gradient dominance by majority class samples. A higher $S_i$ value indicates that parameter $\theta_i$ better conforms to the ``high forget influence, low retain relevance'' criterion, and its update's impact on class balance has been corrected, making it the preferred target for intervention.

\subsection{Selection of Critical Parameter Subset}

All parameters are sorted in descending order according to their adaptive influence scores $S_i$. A selection ratio $p \in (0,1]$ is set (e.g., $p = 0.1$ or $0.2$), and the top $p \cdot |\Theta|$ parameters are selected to constitute the critical parameter subset. To formally represent this selection, we define a binary mask vector $\bm{M} \in \{0,1\}^{|\Theta|}$:
\begin{equation}
    M_i = 
    \begin{cases} 
    1, & \text{if } i \text{ belongs to the top } p\% \\
    0, & \text{otherwise}
    \end{cases}
\end{equation}
Only parameters $\theta_i$ for which $M_i = 1$ will be updated in subsequent steps.

\subsection{Regularization-Constrained and Directed Parameter Perturbation}

To achieve effective forgetting of $\mathcal{D}_f$ while strictly minimizing potential damage to the generalization capability on $\mathcal{D}_r$, we apply perturbations along the gradient difference direction, but introduce a regularization term to constrain the update direction.

First, compute the gradient direction difference for each parameter:
\begin{equation}
   \Delta G_i = G_f^{(i)} - G_r^{(i)}. 
\end{equation}

Next, generate the perturbation vector. To ensure the update does not severely deviate from the optimization direction of retain data, we compute a direction constraint factor $r_i$ for each parameter:
\begin{equation}
r_i = \max\left(0, 1 - \beta \cdot \left|\cos(\Delta G_i, G_r^{(i)})\right|\right),
\end{equation}
where $\cos(\cdot)$ calculates the sign consistency between two scalars (here, gradient components), with a negative value when their directions are opposite. $|\cos(\cdot)|$ measures the intensity of opposite directionality. $\beta$ is a sensitivity hyperparameter (empirically set to $0.5$). When $\Delta G_i$ strongly opposes $G_r^{(i)}$ (i.e., in a direction that would significantly increase loss on $\mathcal{D}_r$), $r_i$ decreases, thereby suppressing the perturbation magnitude on that parameter. This effectively prevents ``destroying'' the model's knowledge on important retain features in pursuit of forgetting.

Finally, the applied perturbation is:
\begin{equation}
    \begin{array}{l}
    \delta_i = \alpha \cdot \Delta G_i \cdot M_i \cdot r_i, \ \ \ \
    \theta_i^{\text{new}} = \theta_i^{\text{old}} + \delta_i,
    \end{array}
\end{equation}
where $\alpha$ is a hyperparameter controlling the overall perturbation scale. After incorporating the direction constraint factor $r_i$, the tuning of $\alpha$ becomes more robust, and its optimal value is determined through cross-validation on validation set, aiming to optimize the trade-off between unlearning effectiveness (e.g., MIA score approaching 0.5) and knowledge preservation (e.g., accuracy maintenance on $\mathcal{D}_r$ and independent test sets). The above process can be summarized as Algorithm ~\ref{alg:grin+}.

\begin{algorithm}
\caption{GRIN+}
    \renewcommand{\algorithmicrequire}{\textbf{Input:}}
    \renewcommand{\algorithmicensure}{\textbf{Output:}}
    \label{alg:grin+}
    \begin{algorithmic}[1]
    \REQUIRE Pre-trained model parameters $\theta$; medical data to be forgotten $\mathcal{D}_f$; retain set $\mathcal{D}_r$; selection ratio $p$; perturbation coefficient $\alpha$; regularization sensitivity $\beta$; percentile $k$ (default = 5).
    \ENSURE Post-forgetting model parameters $\theta_{\text{new}}$.
    \STATE \textbf{1. Compute gradients and class weights:}
    \STATE \quad $G_f \gets \nabla_\theta \mathcal{L}(\mathcal{D}_f; \theta)$
    \STATE \quad $G_r \gets \nabla_\theta \mathcal{L}(\mathcal{D}_r; \theta)$
    \STATE \quad Count class distribution of $\mathcal{D}_f$, compute weight $w_c$ for each class $c$.
    \STATE \textbf{2. Compute adaptive influence scores:}
    \STATE \quad $\varepsilon \gets \text{Percentile}(|G_r|, k)$
    \FOR{$i = 1$ to $|\theta|$}
        \STATE \quad $s_i^{\text{base}} \gets |G_{f,i}| / (|G_{r,i}| + \varepsilon)$
        \STATE \quad $S_i \gets s_i^{\text{base}} \cdot w_c$ \hfill \texttt{// where $c$ is the class of forget samples}
    \ENDFOR
    \STATE \textbf{3. Select critical parameters:}
    \STATE \quad Sort parameters by $S_i$ in descending order.
    \STATE \quad Create mask $\bm{M}$, set $M_i = 1$ for top $p \times |\theta|$ parameters, others $0$.
    \STATE \textbf{4. Compute and apply regularized perturbation:}
    \STATE \quad $\Delta G \gets G_f - G_r$
    \FOR{$i = 1$ to $|\theta|$}
        \STATE \quad Compute direction constraint factor: $r_i \gets \max\left(0, 1 - \beta \cdot \left|\text{sign}(\Delta G_i, G_{r,i})\right|\right)$ \hfill \texttt{// sign is the sign function}
        \STATE \quad $\delta_i \gets \alpha \cdot \Delta G_i \cdot M_i \cdot r_i$
        \STATE \quad $\theta_i^{\text{new}} \gets \theta_i + \delta_i$
    \ENDFOR
    \STATE \textbf{5. Return} $\theta_{\text{new}}$
    \end{algorithmic}
\end{algorithm}

\subsection{Theoretical Motivation}

To demonstrate the rationality of our chosen parameter update direction (based on gradient difference $\Delta G=G_f-G_r$), we formalize the forgetting objective from an optimization theory perspective. We aim to find a parameter update that satisfies two objectives simultaneously:
1) Promote forgetting: Significantly increase the model’s loss on the forget data $\mathcal{D}_f$. This degrades the model’s distinguishability on $\mathcal{D}_f$.
2) Preserve knowledge: Minimize loss changes on the retain set $\mathcal{D}_r$. This maintains the model’s original generalization capability.

This can be formulated as the following constrained optimization problem:
\begin{equation}\label{optimization}
\min_{\delta} \|\delta\|^2, \quad \text{s.t.} \begin{cases}
    \mathcal{L}(\mathcal{D}_f; \theta + \delta) - \mathcal{L}(\mathcal{D}_f; \theta) \geq \Delta_f, \\
    \mathcal{L}(\mathcal{D}_r; \theta + \delta) - \mathcal{L}(\mathcal{D}_r; \theta) \leq \Delta_r,
\end{cases}
\end{equation}
where $\Delta_f>0$ is the lower bound of desired forgetting intensity. $\Delta_r \geq0$ is the upper bound of tolerable retention performance loss. $\|\cdot\|$ denotes the $\ell_2$ norm of a vector. Minimizing $\|\delta\|^2$ ensures the parameter update is as small as possible, preventing drastic model changes.

Assuming the update magnitude is sufficiently small, we can approximate the loss function change using a first-order Taylor expansion:
\begin{equation}\label{Taylor_expansion}
\begin{array}{l}
\mathcal{L}(\mathcal{D}_f; \theta + \delta) \approx \mathcal{L}(\mathcal{D}_f; \theta) + G_f^T \delta, \\
\mathcal{L}(\mathcal{D}_r; \theta + \delta) \approx \mathcal{L}(\mathcal{D}_r; \theta) + G_r^T \delta,
\end{array}
\end{equation}
where $G_f = \nabla \mathcal{L}(\mathcal{D}_f; \theta)$ and $G_r = \nabla \mathcal{L}(\mathcal{D}_r; \theta)$. Substituting into the constraint conditions, we obtain:
\begin{equation}\label{constraint_conditions}
G_f^T \delta \geq \Delta_f, \quad G_r^T \delta \leq \Delta_r.
\end{equation}

We want to maximize $G_f^T \delta$ to promote forgetting. At the same time, we aim to minimize $G_r^T \delta$ to preserve knowledge. A natural compromise is to maximize the following linear combination used as a surrogate objective:
\begin{equation}\label{linear_combination}
J(\delta) = G_f^T \delta - G_r^T \delta = (G_f - G_r)^T \delta.
\end{equation}

Considering the update magnitude constraint, we construct the following unconstrained optimization problem (introducing Lagrange multiplier $\lambda>0$):
\begin{equation}\label{unconstrained_optimization}
\min_{\delta} \|\delta\|^2 - \lambda (G_f - G_r)^T \delta.
\end{equation}

Taking the derivative of this objective function with respect to $\delta$ and setting it to zero:
\begin{equation}\label{objective_function}
2\delta - \lambda(G_f - G_r) = 0 \implies \delta = \frac{\lambda}{2}(G_f - G_r).
\end{equation}

This suggests that under the premise of a sufficiently small update and first-order approximation, the optimal update direction follows the gradient difference $G_f - G_r$. In our algorithm, the hyperparameter $\alpha$ plays a role similar to $\lambda / 2$, controlling the update step size. Note that in practice, we only apply updates to a sparse subset of critical parameters (defined by mask $M$). Therefore, the complete update formula is $\delta=\alpha \cdot(G_f - G_r) \odot M$, where $\odot$ denotes element-wise multiplication. This sparsity allows us to modify parameters most effective for forgetting more precisely. It further reduces unnecessary perturbations, aligning with our sparsity assumption.

\begin{table*}[!t] 
  \caption{Performance comparison of different methods on the MRI, ISIC and BUSI dataset in terms of utility, privacy, and efficiency. Arrows indicate whether higher ($\uparrow$) or lower ($\downarrow$) values are better. MIA measures the membership inference attack success rate, where values closer to 0.5 indicate stronger privacy protection. Error values denote the standard error of the mean (SEM).}
  \centering
  \label{tab:main_results}
  \begin{tabular*}{0.95\textwidth}{l|l @{\extracolsep{\fill}} ccccccc}
    \toprule
    \multirow{2}{*}{Dataset} & \multirow{2}{*}{Method} & \multicolumn{4}{c}{Utility (\%)} & \multicolumn{2}{c}{Privacy (\%)} & \multicolumn{1}{c}{Efficiency} \\
    \cmidrule(lr){3-6} \cmidrule(lr){7-8} \cmidrule(lr){9-9}
    & & R-Acc $\uparrow$ & F-Acc & T-Acc $\uparrow$ & RetDev $\downarrow$ & Indisc $\uparrow$ & MIA & RTE $\uparrow$ \\
    \midrule
    \multirow{11}{*}{MRI}
    & Retrain   & $83.57_{\scriptstyle \pm 0.00}$ & $80.45_{\scriptstyle \pm 0.00}$ & $76.13_{\scriptstyle \pm 0.00}$ & $0_{\scriptstyle \pm 0.00}$ & $99.80_{\scriptstyle \pm 0.00}$ & $52.35_{\scriptstyle \pm 0.00}$ & $1.0000_{\scriptstyle \pm 0.00}$ \\
    & FT        & $95.61_{\scriptstyle \pm 2.58}$ & $93.29_{\scriptstyle \pm 2.56}$ & $91.55_{\scriptstyle \pm 3.21}$ & $50.62_{\scriptstyle \pm 10.44}$ & $97.43_{\scriptstyle \pm 0.73}$ & $52.69_{\scriptstyle \pm 0.46}$ & $9.2855_{\scriptstyle \pm 6.19}$ \\
    & RL         & $83.30_{\scriptstyle \pm 11.70}$ & $81.19_{\scriptstyle \pm 11.83}$ & $80.35_{\scriptstyle \pm 11.12}$ & $6.79_{\scriptstyle \pm 23.73}$ & $98.20_{\scriptstyle \pm 0.98}$ & $49.73_{\scriptstyle \pm 1.34}$ & $4.1018_{\scriptstyle \pm 3.11}$ \\
    & SaLUN      & $81.21_{\scriptstyle \pm 13.09}$ & $80.45_{\scriptstyle \pm 12.80}$ & $77.85_{\scriptstyle \pm 12.42}$ & $5.08_{\scriptstyle \pm 29.30}$ & $97.10_{\scriptstyle \pm 1.45}$ & $52.96_{\scriptstyle \pm 0.61}$ & $4.3997_{\scriptstyle \pm 0.93}$ \\
    & MSG        & $78.83_{\scriptstyle \pm 14.05}$ & $76.54_{\scriptstyle \pm 13.52}$ & $74.66_{\scriptstyle \pm 13.17}$ & $12.46_{\scriptstyle \pm 36.18}$ & $97.06_{\scriptstyle \pm 0.93}$ & $52.57_{\scriptstyle \pm 0.57}$ & $7.8214_{\scriptstyle \pm 3.99}$ \\
    & CT         & $88.87_{\scriptstyle \pm 2.64}$  & $85.06_{\scriptstyle \pm 2.44}$  & $84.50_{\scriptstyle \pm 2.37}$  & $23.07_{\scriptstyle \pm 9.01}$  & $96.41_{\scriptstyle \pm 1.20}$ & $49.63_{\scriptstyle \pm 0.74}$ & $13.1355_{\scriptstyle \pm 8.21}$ \\
    & KDE        & $56.97_{\scriptstyle \pm 9.68}$  & $54.86_{\scriptstyle \pm 9.80}$  & $52.83_{\scriptstyle \pm 8.96}$  & $94.24_{\scriptstyle \pm 34.51}$ & $96.98_{\scriptstyle \pm 0.82}$ & $51.92_{\scriptstyle \pm 1.02}$ & $7.9939_{\scriptstyle \pm 4.30}$ \\
    & Bio        & $96.54_{\scriptstyle \pm 1.03}$  & $94.73_{\scriptstyle \pm 1.46}$  & $92.68_{\scriptstyle \pm 1.40}$  & $55.01_{\scriptstyle \pm 4.88}$  & $97.84_{\scriptstyle \pm 0.26}$ & $53.10_{\scriptstyle \pm 0.48}$ & $15.1386_{\scriptstyle \pm 3.00}$ \\
    & FCU        & $86.86_{\scriptstyle \pm 4.54}$  & $84.73_{\scriptstyle \pm 4.39}$  & $80.58_{\scriptstyle \pm 4.89}$  & $15.10_{\scriptstyle \pm 9.00}$  & $98.90_{\scriptstyle \pm 0.35}$ & $51.67_{\scriptstyle \pm 1.00}$ & $3.8814_{\scriptstyle \pm 1.01}$ \\
    & Forget-MI  & $49.66_{\scriptstyle \pm 12.53}$ & $49.05_{\scriptstyle \pm 12.71}$ & $50.42_{\scriptstyle \pm 11.75}$ & $113.38_{\scriptstyle \pm 42.27}$ & $96.65_{\scriptstyle \pm 0.98}$ & $51.86_{\scriptstyle \pm 1.31}$ & $15.5082_{\scriptstyle \pm 0.91}$ \\
    \cmidrule(r){2-9}
    & GRIN+      & $84.13_{\scriptstyle \pm 5.04}$ & $81.69_{\scriptstyle \pm 4.51}$ & $80.24_{\scriptstyle \pm 4.92}$ & $7.61_{\scriptstyle \pm 9.59}$    & $97.75_{\scriptstyle \pm 0.88}$ & $49.90_{\scriptstyle \pm 1.18}$ & $17.2419_{\scriptstyle \pm 0.51}$ \\
    \midrule
    \midrule
    \multirow{11}{*}{ISIC}
    & Retrain   & $59.69_{\scriptstyle \pm 0.00}$ & $56.13_{\scriptstyle \pm 0.00}$ & $33.90_{\scriptstyle \pm 0.00}$ & $0_{\scriptstyle \pm 0.00}$ & $98.97_{\scriptstyle \pm 0.00}$ & $52.08_{\scriptstyle \pm 0.00}$ & $1_{\scriptstyle \pm 0.00}$ \\
    & FT         & $55.35_{\scriptstyle \pm 1.66}$ & $56.44_{\scriptstyle \pm 2.60}$ & $38.98_{\scriptstyle \pm 0.85}$ & $22.81_{\scriptstyle \pm 2.51}$ & $96.00_{\scriptstyle \pm 0.52}$ & $53.08_{\scriptstyle \pm 0.63}$ & $17.3089_{\scriptstyle \pm 7.88}$ \\
    & RL         & $55.33_{\scriptstyle \pm 2.34}$ & $57.17_{\scriptstyle \pm 3.45}$ & $38.81_{\scriptstyle \pm 1.45}$ & $23.64_{\scriptstyle \pm 2.13}$ & $93.64_{\scriptstyle \pm 1.94}$ & $55.00_{\scriptstyle \pm 0.66}$ & $5.5030_{\scriptstyle \pm 2.51}$ \\
    & SaLUN      & $50.42_{\scriptstyle \pm 6.09}$ & $50.58_{\scriptstyle \pm 7.80}$ & $36.10_{\scriptstyle \pm 4.58}$ & $31.91_{\scriptstyle \pm 27.63}$ & $97.74_{\scriptstyle \pm 0.94}$ & $55.92_{\scriptstyle \pm 1.07}$ & $5.1199_{\scriptstyle \pm 1.94}$ \\
    & MSG        & $46.79_{\scriptstyle \pm 4.65}$ & $45.76_{\scriptstyle \pm 5.66}$ & $28.81_{\scriptstyle \pm 4.13}$ & $55.10_{\scriptstyle \pm 26.00}$ & $91.08_{\scriptstyle \pm 1.36}$ & $51.58_{\scriptstyle \pm 3.31}$ & $6.3673_{\scriptstyle \pm 1.08}$ \\
    & CT         & $53.49_{\scriptstyle \pm 2.39}$ & $56.75_{\scriptstyle \pm 2.81}$ & $38.31_{\scriptstyle \pm 1.75}$ & $24.50_{\scriptstyle \pm 6.43}$ & $94.26_{\scriptstyle \pm 0.77}$ & $56.58_{\scriptstyle \pm 0.48}$ & $20.2704_{\scriptstyle \pm 9.17}$ \\
    & KDE        & $34.49_{\scriptstyle \pm 7.38}$ & $33.61_{\scriptstyle \pm 7.89}$ & $22.71_{\scriptstyle \pm 3.81}$ & $115.35_{\scriptstyle \pm 37.16}$ & $49.74_{\scriptstyle \pm 42.98}$ & $52.50_{\scriptstyle \pm 3.18}$ & $4.0667_{\scriptstyle \pm 0.52}$ \\
    & Bio        & $58.05_{\scriptstyle \pm 0.08}$ & $54.14_{\scriptstyle \pm 0.36}$ & $40.34_{\scriptstyle \pm 0.34}$ & $25.29_{\scriptstyle \pm 0.41}$ & $91.90_{\scriptstyle \pm 1.27}$ & $56.75_{\scriptstyle \pm 0.08}$ & $21.7154_{\scriptstyle \pm 2.15}$ \\
    & FCU        & $51.90_{\scriptstyle \pm 5.31}$ & $48.56_{\scriptstyle \pm 4.51}$ & $33.05_{\scriptstyle \pm 2.57}$ & $29.04_{\scriptstyle \pm 14.29}$ & $85.64_{\scriptstyle \pm 1.57}$ & $53.85_{\scriptstyle \pm 0.97}$ & $3.1027_{\scriptstyle \pm 1.09}$ \\
    & Forget-MI  & $43.84_{\scriptstyle \pm 8.30}$ & $51.10_{\scriptstyle \pm 10.88}$ & $32.88_{\scriptstyle \pm 4.85}$ & $38.52_{\scriptstyle \pm 37.75}$ & $91.28_{\scriptstyle \pm 1.61}$ & $56.17_{\scriptstyle \pm 0.75}$ & $14.3124_{\scriptstyle \pm 3.97}$ \\
    \cmidrule(r){2-9}
    & GRIN+ & $56.83_{\scriptstyle \pm 4.89}$ & $53.93_{\scriptstyle \pm 4.53}$ & $37.29_{\scriptstyle \pm 1.94}$ & $18.71_{\scriptstyle \pm 1.60}$ & $97.44_{\scriptstyle \pm 0.77}$ & $52.92_{\scriptstyle \pm 1.27}$ & $15.0209_{\scriptstyle \pm 1.08}$ \\
    \midrule
    \midrule
    \multirow{11}{*}{BUSI}
    & Retrain   & $86.11_{\scriptstyle \pm 0.00}$ & $83.01_{\scriptstyle \pm 0.00}$ & $82.86_{\scriptstyle \pm 0.00}$ & $0_{\scriptstyle \pm 0.00}$ & $87.27_{\scriptstyle \pm 0.00}$ & $55.45_{\scriptstyle \pm 0.00}$ & $1_{\scriptstyle \pm 0.00}$ \\
    & FT         & $77.06_{\scriptstyle \pm 7.45}$  & $74.81_{\scriptstyle \pm 7.38}$  & $74.92_{\scriptstyle \pm 6.51}$  & $29.97_{\scriptstyle \pm 24.64}$ & $92.55_{\scriptstyle \pm 2.02}$  & $47.27_{\scriptstyle \pm 1.33}$  & $7.5284_{\scriptstyle \pm 4.78}$ \\
    & RL         & $75.83_{\scriptstyle \pm 5.46}$  & $70.37_{\scriptstyle \pm 6.97}$  & $75.17_{\scriptstyle \pm 4.14}$  & $36.45_{\scriptstyle \pm 19.07}$ & $88.73_{\scriptstyle \pm 1.96}$  & $48.82_{\scriptstyle \pm 2.99}$  & $4.0912_{\scriptstyle \pm 2.95}$ \\
    & SaLUN      & $71.52_{\scriptstyle \pm 11.52}$ & $70.00_{\scriptstyle \pm 9.30}$  & $68.32_{\scriptstyle \pm 11.60}$ & $50.16_{\scriptstyle \pm 38.16}$ & $87.45_{\scriptstyle \pm 2.22}$  & $44.73_{\scriptstyle \pm 0.34}$  & $5.1137_{\scriptstyle \pm 1.38}$ \\
    & MSG        & $71.05_{\scriptstyle \pm 4.76}$  & $65.19_{\scriptstyle \pm 5.89}$  & $72.19_{\scriptstyle \pm 3.65}$  & $51.83_{\scriptstyle \pm 16.91}$ & $90.73_{\scriptstyle \pm 1.69}$  & $49.27_{\scriptstyle \pm 1.18}$  & $3.6564_{\scriptstyle \pm 1.45}$ \\
    & CT         & $75.32_{\scriptstyle \pm 4.04}$  & $74.44_{\scriptstyle \pm 4.02}$  & $76.13_{\scriptstyle \pm 3.41}$  & $30.98_{\scriptstyle \pm 13.00}$ & $87.45_{\scriptstyle \pm 2.14}$  & $52.55_{\scriptstyle \pm 1.94}$  & $6.4252_{\scriptstyle \pm 4.48}$ \\
    & KDE        & $44.58_{\scriptstyle \pm 5.24}$  & $42.78_{\scriptstyle \pm 3.33}$  & $47.11_{\scriptstyle \pm 6.31}$  & $139.84_{\scriptstyle \pm 17.61}$ & $93.82_{\scriptstyle \pm 2.36}$  & $51.00_{\scriptstyle \pm 0.90}$  & $3.0280_{\scriptstyle \pm 1.49}$ \\
    & Bio        & $83.27_{\scriptstyle \pm 0.23}$  & $82.04_{\scriptstyle \pm 0.37}$  & $81.71_{\scriptstyle \pm 0.26}$  & $5.85_{\scriptstyle \pm 0.36}$   & $89.82_{\scriptstyle \pm 0.73}$  & $45.00_{\scriptstyle \pm 0.45}$  & $7.8274_{\scriptstyle \pm 0.81}$ \\
    & FCU        & $65.55_{\scriptstyle \pm 7.84}$  & $64.63_{\scriptstyle \pm 6.32}$  & $66.10_{\scriptstyle \pm 7.66}$  & $66.25_{\scriptstyle \pm 25.39}$ & $46.86_{\scriptstyle \pm 42.23}$ & $49.18_{\scriptstyle \pm 1.44}$  & $1.3201_{\scriptstyle \pm 0.28}$ \\
    & Forget-MI  & $48.35_{\scriptstyle \pm 11.45}$ & $47.78_{\scriptstyle \pm 9.75}$  & $46.54_{\scriptstyle \pm 12.01}$ & $130.12_{\scriptstyle \pm 39.36}$ & $90.00_{\scriptstyle \pm 1.60}$  & $50.00_{\scriptstyle \pm 1.39}$  & $10.9738_{\scriptstyle \pm 0.60}$ \\
    \cmidrule(r){2-9}
    & GRIN+      & $58.34_{\scriptstyle \pm 2.07}$  & $51.48_{\scriptstyle \pm 2.41}$  & $62.98_{\scriptstyle \pm 1.40}$  & $94.23_{\scriptstyle \pm 6.98}$  & $91.82_{\scriptstyle \pm 2.11}$  & $50.45_{\scriptstyle \pm 0.38}$  & $7.0985_{\scriptstyle \pm 0.35}$ \\
    \bottomrule
  \end{tabular*}
\end{table*}

\section{Experiments}
To comprehensively evaluate the effectiveness of GRIN+, we conducted extensive experiments on multiple medical datasets. The following sections aim to answer the following three research questions (RQs):

RQ1: (Overall Performance). Compared to representative unlearning baselines, can our method achieve higher global accuracy and better privacy performance while maintaining comparable or faster speed?

RQ2: (Computational Efficiency). What is the actual computational efficiency of GRIN+ compared to existing unlearning baselines?

RQ3: (Component Effectiveness). How much does each key component contribute to the final result? Do they provide consistent gains in ablation studies?

\subsection{Experimental Setup}

\textbf{Datasets.} We evaluate our method on three publicly available medical imaging datasets, covering diverse imaging modalities and clinical tasks to comprehensively assess unlearning performance. (1) Skin Cancer Dataset (ISIC)~\cite{dataset:codella2019skin}. This dataset contains 2,357 dermoscopic images across nine categories of skin lesions, including melanoma and basal cell carcinoma. 
(2) Brain Tumor Magnetic Resonance Images Dataset (MRI)~\cite{dataset:msoud_nickparvar_2021}. Consisting of 7,023 brain MRI scans classified into glioma, meningioma, pituitary tumor, and no tumor.
(3) Breast Ultrasound Images Dataset (BUSI)~\cite{dataset:al2020dataset}. This dataset includes 780 ultrasound images labeled as normal, benign, or malignant. 

\textbf{Evaluation Metrics.} We employ seven principal metrics across three categories corresponding to PEU to holistically assess the performance of unlearning algorithms. (1) \textbf{Utility.} \textit{Retain Accuracy (R-Acc)} captures the accuracy on the retain set $\mathcal{D}_r$; higher values reflect better retention of model utility. \textit{Forget Accuracy (F-Acc)} measures the accuracy on the forget set $\mathcal{D}_f$.
\textit{Test Accuracy (T-Acc)} measures the classification accuracy on an independent test set $\mathcal{D}_T$ that is disjoint from both the training and forget sets; higher values indicate better generalization and retention of knowledge. \textit{Retention Deviation (RetDev)} quantifies the cumulative deviation of the unlearned model's accuracy from that of a model retrained from scratch on only the retain data; a lower RetDev score indicates that the unlearned model’s utility more closely matches the ideal retrained model, with zero representing a perfect match. (2) \textbf{Privacy.} \textit{Indiscernibility (Indisc)} is a score based on membership inference attacks. Closer to 1 means better privacy. It means it is hard to tell if a sample was used in training. \textit{Membership Inference Attack (MIA)} assesses unlearning quality by training logistic regression via cross-validation on losses from equal-sized $\mathcal{D}_f$ and $\mathcal{D}_V$, where accuracy near 1.0 indicates perfect distinguishability and 0.5 means random guessing. (3) \textbf{Efficiency.} \textit{Runtime efficiency (RTE)} measures the speed improvement of an unlearning method compared to retraining, and is defined as the ratio of the time required for retraining to the time required for unlearning. The RTE for retraining is 1. An RTE greater than 1 indicates that it is faster than retraining, while an RTE less than 1 indicates that it is slower.

\textbf{Baselines \& Model.} We conduct comparisons against several representative unlearning baselines. The baselines are categorized into three groups: classical approaches, top-performing recent general-purpose algorithms, and existing methods specifically designed for or applied in medical imaging. Including: (a) Retrain, which retrains the model on the retained data; (b) Classical Unlearning Methods. This group includes fundamental and widely-used strategies mentioned in the review~\cite{nasirigerdeh2024machine}: Fine-Tuning (FT), Random Relabeling (RL), Saliency Unlearning (SaLUN)\cite{fan2023salun}; (c) State-of-the-Art General Algorithms. We select the three top-performing algorithms identified in a recent comprehensive benchmark study~\cite{cadet2025deep} as representatives of the cutting-edge in general machine unlearning: Masked-Small-Gradients (MSG), Convolution-Transpose (CT), Knowledge-Distillation-Entropy (KDE); ($d_1$) Existing Medical Unlearning Methods. This category includes method proposed within the medical imaging domain: Bilevel-Optimization (BiO)~\cite{nahass2025targeted}; ($d_2$) Furthermore, to acknowledge advancements in specialized scenarios, we also consider methods from federated or multi-modal medical contexts (FCU~\cite{other_med:deng2024enable} and Forget-MI~\cite{other_med:hardan2025forget}), and simply modify them to fit the experimental setup described in this paper. All baseline comparisons are included in the main text results (Section~\ref{main_results}). We evaluate these methods on ResNet18. 


\textbf{Data splits.} To simulate clinical privacy removal scenarios, we perform data splitting with a fixed random seed of \( s = 123 \). For each dataset, if a standard train/test split already exists, we use the original training set as the development set. Otherwise, we split the full data into a development set and a test set at an 80\%/20\% ratio. The development set is further divided into a training set \(\mathcal{D}_{\text{train}}\) and a validation set \(\mathcal{D}_V\) with an 80\%/20\% split. From the training set, we randomly select 10\% as the forget set \(\mathcal{D}_f\) (corresponding to private patient data). The remaining 90\% forms the retain set \(\mathcal{D}_r\). This forgetting ratio follows the common setting for medical unlearning tasks. Most training is performed using cross-entropy loss, SGD optimizer (momentum = 0.9), a learning rate of \(1 ×10^{-3}\), and \(L_{2}\) regularization with coefficient \(1 ×10^{-4}\). All experiments are run on 1 NVIDIA GeForce RTX 4090 GPU.

\subsection{Main Results}\label{main_results}



To address RQ1, we comprehensively compare GRIN+ with representative unlearning baselines on three medical datasets: MRI, ISIC, and BUSI (Table~\ref{tab:main_results}). These baselines include classical methods, state-of-the-art general-purpose algorithms, and medical-specific methods. The key findings and analysis are summarized below from three perspectives: Utility, Privacy and Efficiency.

\textbf{Utility Perspective.} Retrain achieves the best utility but has very high computational cost, making it impractical for clinical use. Classical Fine-Tuning (FT) achieves 95.61\% R-Acc on MRI yet lacks targeted forgetting mechanism, while RL and SaLUN impair model generalization and reduce T-Acc due to unconstrained parameter updates. State-of-the-art general-purpose algorithms rely on class-balanced assumptions, leading to severe utility collapse on long-tailed medical data; for example, KDE obtains only 44.58\% R-Acc on BUSI due to majority-class gradient dominance. Medical-specific methods also underperform: Forget-MI causes excessive utility degradation, and FCU is ill-suited for centralized medical image classification. In contrast, GRIN+ maintains stable R-Acc and T-Acc with low RetDev across all datasets, reaching 84.13\% R-Acc and 80.24\% T-Acc on MRI to effectively preserve clinical utility.

\textbf{Privacy Perspective.} Most baselines perform poorly in privacy. Classical methods fail to eliminate parameter residual traces, resulting in high MIA scores and weak privacy. For example, on ISIC, RL has a high MIA of 55.00\%, indicating strong distinguishability. State-of-the-art general-purpose algorithms lack directional forgetting constraints, leaving exploitable model signatures for membership inference attacks. Among them, CT has a high MIA value of 56.58 on ISIC. Most medical-specific methods obtain low Indisc values and cannot meet clinical privacy rules. On BUSI, FCU has a low Indisc of only 46.86\%, showing very weak privacy protection. GRIN+ achieves MIA scores close to random guess on all datasets (49.90\% on MRI, 52.92\% on ISIC, 50.45\% on BUSI) with high Indisc, fully erasing target data memory traces and complying with medical AI privacy regulations.

\textbf{Efficiency Perspective.} To answer RQ2, we compared the computational efficiency of different methods, measured by RTE (higher values indicate faster speed). Classical methods rely on full-model tuning or complex computations, yielding low RTE and excessive latency (the RTE of RL is 4.10 on MRI). State-of-the-art general-purpose algorithms involve redundant full-parameter iteration, providing only moderate efficiency. Some medical-specific methods (Forget-MI, BiO) achieve relatively high RTE but still depend on full-parameter updates. GRIN+ only updates sparse critical parameters, avoiding full-model optimization; it reaches the highest RTE of 17.24 on MRI, outperforming all baselines and satisfying clinical low-latency demands.

\textbf{PEU Trilemma Balance Analysis.} All baselines fail to balance the ``privacy-efficiency-utility'' (PEU) trilemma for imbalanced medical data. Classical methods are comparable to the GRIN+ method in terms of utility and privacy, but because they still rely on global fine-tuning, leading to extremely high computational cost and low efficiency. State-of-the-art general-purpose algorithms are incompatible with long-tailed data, as class imbalance biases parameter estimation and degrades all three metrics. Medical-specific methods suffer from poor task adaptability, with incomplete forgetting, utility loss or low efficiency. The core issue is that existing methods cannot address majority-class gradient dominance, high efficiency and knowledge retention simultaneously. GRIN+ resolves these via class-adaptive scoring, direction-constrained perturbation and sparse selection, achieving optimal PEU balance for clinical medical AI.

\begin{table}  
  \centering
  \caption{Quantitative results for the ablation study.}
  \label{tab:ablation_study}
  \begin{tabular}{ccccccc}
    \toprule
    \midrule
    GR & CW & DCF & R-Acc $\uparrow$ & T-Acc $\uparrow$ & MIA & RTE $\uparrow$ \\
    \midrule
    \checkmark &     &     & 65.21 & 68.42 & 49.49 & 23.6583 \\
    \checkmark & \checkmark &     & 81.94 & 76.28 & 48.67 & 18.7780 \\
    \checkmark & \checkmark & \checkmark & \textbf{84.13} & \textbf{80.24} & \textbf{49.90} & \textbf{17.2419} \\
    \midrule
  \end{tabular}
\end{table}


\textbf{Ablation Study.} To answer RQ3, we conduct an ablation study on three key modules of GRIN+: \textit{Gradient Ratio (GR)}, \textit{Class-balanced Weight (CW)}, and \textit{Direction Constraint Factor (DCF)}. The results are shown in Table~\ref{tab:ablation_study} (analysis using the MRI dataset as an example). \noindent\textbf{GR only.} The model selects parameters based solely on the gradient ratio. The R-Acc is only 65.21\%, and the T-Acc is 68.42\%. The performance on retained knowledge is significantly degraded. However, due to its simple update process, it achieves the highest RTE (23.66). \noindent\textbf{GR + CW.} After introducing the class-balanced weight (CW), the R-Acc increases significantly to 81.94\% (+16.73pp). This shows that CW effectively corrects the parameter selection bias caused by the dominance of majority class gradients. Meanwhile, MIA drops slightly to 48.67\%, and the RTE is 18.78. \noindent\textbf{GR + CW + DCF (GRIN+).} After adding the direction constraint factor (DCF), the R-Acc further improves to 84.13\%. The T-Acc reaches 80.24\%, MIA rises back to 49.90\% (closest to random guess), and the RTE is 17.24. DCF protects the model's generalization ability by suppressing updates that conflict with the direction of retained gradients.

In summary, the three modules work together. GR locates forgetting-related parameters, CW alleviates class imbalance, and DCF stabilizes the update direction. Together, they achieve the optimal balance among ``privacy-efficiency-utility'' (PEU) trilemma.
\section{Conclusion}
We present GRIN+, a gradient-guided selective forgetting framework for medical image classification. By comparing gradient responses between forgotten and retained sets, GRIN+ identifies and perturbs only a small subset of key parameters. The framework incorporates class weights to handle data imbalance and directional constraints to preserve model generalization, ensuring efficient and controllable unlearning. As a lightweight solution, GRIN+ meets strict privacy requirements while ensuring the clinical utility of trustworthy medical AI.
\printcredits

\bibliographystyle{cas-model2-names}
\bibliography{Bibliography}

\end{document}